\documentclass[sn-mathphys]{sn-jnl}

\usepackage{graphicx}
\usepackage{amsmath}
\usepackage{array}
\usepackage{booktabs}
\usepackage{multirow}
\usepackage{caption}
\usepackage{subcaption}

\jyear{2021}%

\theoremstyle{thmstyleone}%
\theoremstyle{thmstyletwo}%

\theoremstyle{thmstylethree}%

\begin{document}

\title[Towards Transcervical US Image Guidance  for TORS]{Towards Transcervical Ultrasound Image Guidance  for Transoral Robotic Surgery}


\author*[1]{\fnm{Wanwen} \sur{Chen}}\email{wanwenc@ece.ubc.ca}
\author[1]{\fnm{Megha} \sur{Kalia}}
\author[1]{\fnm{Qi} \sur{Zeng}}
\author[2]{\fnm{Emily H.T.} \sur{Pang}}
\author[1]{\fnm{Razeyeh} \sur{Bagherinasab}}
\author[3]{\fnm{Thomas D.} \sur{Milner}}
\author[2]{\fnm{Farahna} \sur{Sabiq}}
\author[3]{\fnm{Eitan} \sur{Prisman}}
\author[1]{\fnm{Septimiu E.} \sur{Salcudean}}

\affil*[1]{\orgdiv{Department of Electrical and Computer Engineering}, \orgname{The University of British Columbia}, \orgaddress{\city{Vancouver},  \state{BC}, \country{Canada}}}
\affil[2]{\orgdiv{Department of Radiology}, \orgname{Vancouver General Hospital}, \orgaddress{\city{Vancouver},  \state{BC}, \country{Canada}}}
\affil[3]{\orgdiv{Department of Otolaryngology}, \orgname{Vancouver General Hospital}, \orgaddress{\city{Vancouver},  \state{BC}, \country{Canada}}}



\abstract{
\textbf{Purpose:} Trans-oral robotic surgery (TORS) using the da Vinci surgical robot is a new minimally-invasive surgery method to treat oropharyngeal tumors, but it is a challenging operation. Augmented reality (AR) based on intra-operative ultrasound (US) has the potential to enhance the visualization of the anatomy and cancerous tumors to provide additional tools for decision-making in surgery.

\textbf{Methods:} We propose the first complete pipeline for MRI-US-robot-patient registration for US-guided AR system for TORS, with the transducer placed on the neck for a transcervical view. 
Firstly, we perform a novel MRI-to-transcervical 3D US registration study, comprising (i) preoperative MRI to preoperative US registration, and (ii) preoperative to intraoperative US registration to account for tissue deformation due to retraction. 
Secondly, we develop a US-robot calibration method with an optical tracker and demonstrate its use in an AR system that displays anatomy models in the surgeon's console in real-time. 

\textbf{Results:} 
Our AR system achieves a projection error from the US to the stereo cameras of 27.14 and 26.03 pixels (image is 540$\times$960) in a water bath experiment.
The average target registration error (TRE) for MRI to 3D US is 8.90 mm for the 3D US transducer and 5.85 mm for freehand 3D US, and the TRE for pre-intra operative US registraion is 7.90 mm.

\textbf{Conclusion:} 
We demonstrate the feasibility of each component of our proof-of-concept transcervical US-guided AR system for TORS.
Our results show that trans-cervical 3D US is a promising technique for TORS image guidance. }


\maketitle

\section{Introduction}\label{sec1}

    TORS is a new minimally invasive technique that allows surgeons to remove oropharyngeal cancers with better outcomes than with conventional treatment~\cite{moore2012transoral}. 
TORS, however, is challenging because surgeons need to work 
    through the patient's mouth without haptic feedback, requiring a profound knowledge of the oropharynx anatomy~\cite{moore2012transoral}. 
    Image guidance has the potential to improve TORS through anatomy and cancer visualization to optimize the resection margin and to reduce the risk of postoperative hemorrhage.
    
    Current TORS image guidance approaches mostly visualize preoperative computed tomography (CT) or magnetic resonance imaging (MRI)~\cite{desai2008transoral,pratt2018transoral,chan2020augmented}. 
    CT is the most common intraoperative imaging modality. Comparisons between preoperative and intraoperative CT show that the former is insufficient to visualize the deformed anatomy, while the latter can improve fiducial localization accuracy and task efficiency~\cite{ma2017intraoperative,kahng2019improving}. 
    A tri-planar CT-based surgical navigation system is presented in~\cite{shi2022surgical}.
    Deformable registration from preoperative images to cone-beam CT to augment anatomy structures and surgical planning in stereoscopic video is presented in~\cite{liu2015augmented}.
    However, intra-operative CT introduces unwanted radiation risk to clinicians and takes up space in the OR, so it is less practical than US.
    
    US can provide real-time intra-operative imaging for TORS without radiation. 
    In previous research, the US transducer was placed in the patient's oral cavity, therefore
    TORS robot tools must be removed~\cite{shen2019framework,goepfert2015trans,clayburgh2016intraoperative} 
    or a miniaturized US transducer with limited scanning range must be used~\cite{liu2020abstract,green2020integrated,chang2021real}.
    Transcervical US, which places the US transducer outside the patient’s neck, can provide continuous US imaging during TORS. 
    Transcervical 3D US has been used in neck fine-needle puncture~\cite{schipper2014ultrasound} and neck tumor dissection~\cite{snyder2014neck}, 
    as well as in oral cancer diagnosis~\cite{klimek1998three,rebol2008volume,hong2015efficiency}, but its feasibility in TORS has not been studied.  
    
    While US is accessible and safe, US images are harder to interpret than CT or MRI. 
    US-to-preoperative image registration can combine the benefits of different image modalities to assist surgical planning. In particular, as a pre-operative imaging modality, MRI provides high soft-tissue contrast and cancer imaging. 
    MRI-US registration and real-time US have been used in other surgeries such as prostatectomy~\cite{mathur2019feasibility,kalia2021preclinical}. We hypothesize that they can be applied to TORS, but the feasibility of MRI-3D US registration in the oropharynx is not well studied. 
    Chen {\em et al.}~\cite{chen2022Feasibility} conducted a feasibility study of MRI-3D US registration for TORS, showing that the 3D images acquired by a low-frequency 3D transducer might not have sufficient resolution for image registration.

    We propose a novel AR approach for TORS utilizing transcervical 3D US and evaluate its feasibility. To the best of our knowledge, this is the first work: (1) demonstrating the feasibility of MRI to freehand 3D US registration in the oropharynx; (2) evaluating preoperative to intraoperative US registration; (3) developing a proof-of-concept transcervical US-guided TORS AR system. 
    
\section{Materials and Methods}\label{sec:system}

\begin{figure}
            \centering
            \includegraphics[width=0.9\textwidth]{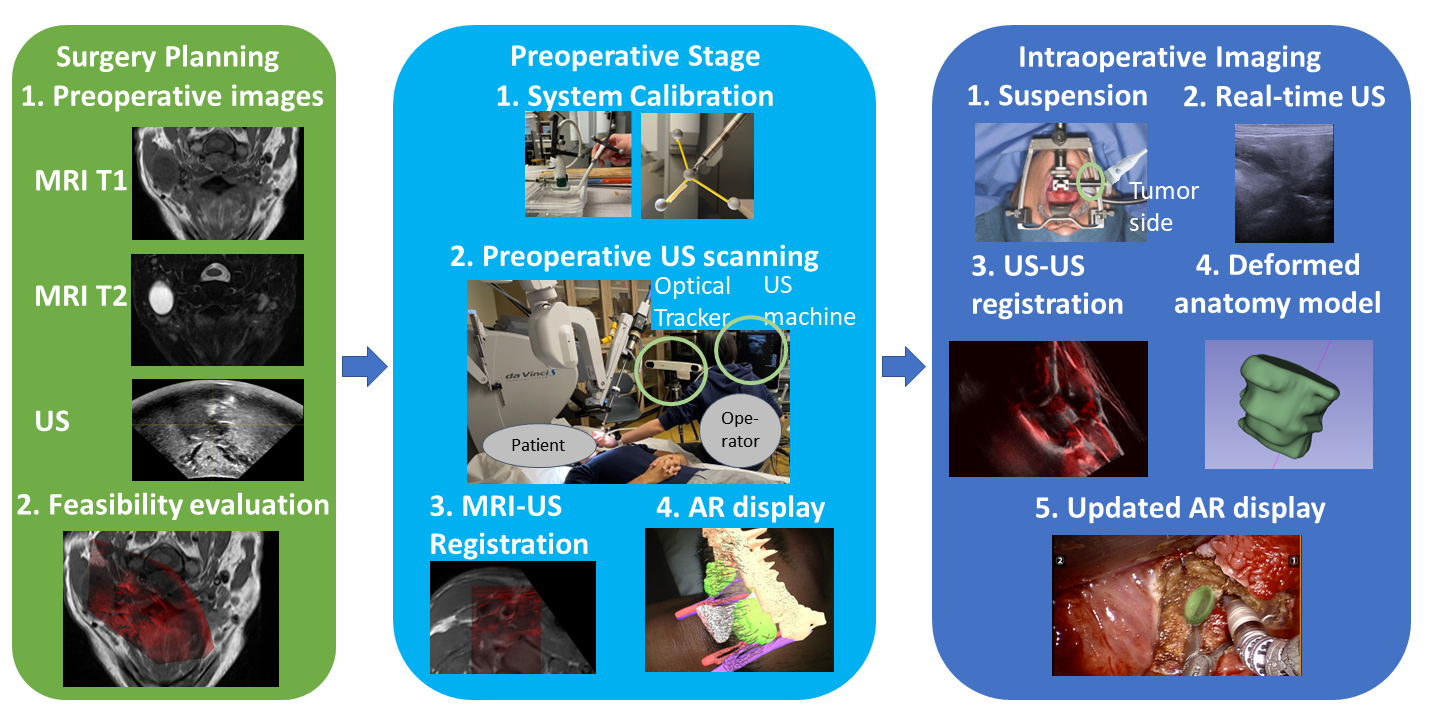}
            \caption{The proposed workflow of the AR-guided TORS.}
            \label{fig:surgical_worflow}
        \end{figure}

    Fig.~\ref{fig:surgical_worflow} shows the proposed US-guided AR workflow. In the planning stage, preoperative MRI and US are collected for surgical planning. Just before  surgery, the AR system is calibrated and a preoperative freehand US scanning and MRI-US registration are carried out for an initial AR display of the anatomy model. During surgery, the patient's tongue is retracted, and the surgeon can use the US transducer to scan the neck on the tumor side. A US-to-US pre-intra-operative registration generates the deformation field to update the anatomy model in AR. In this preliminary study, we evaluate the feasibility of system calibration, MRI-US registration and US-US registration.
    \subsection{System Hardware}\label{sec:system-hardware}
\begin{figure}
            \centering
            \includegraphics[width=0.9\linewidth]{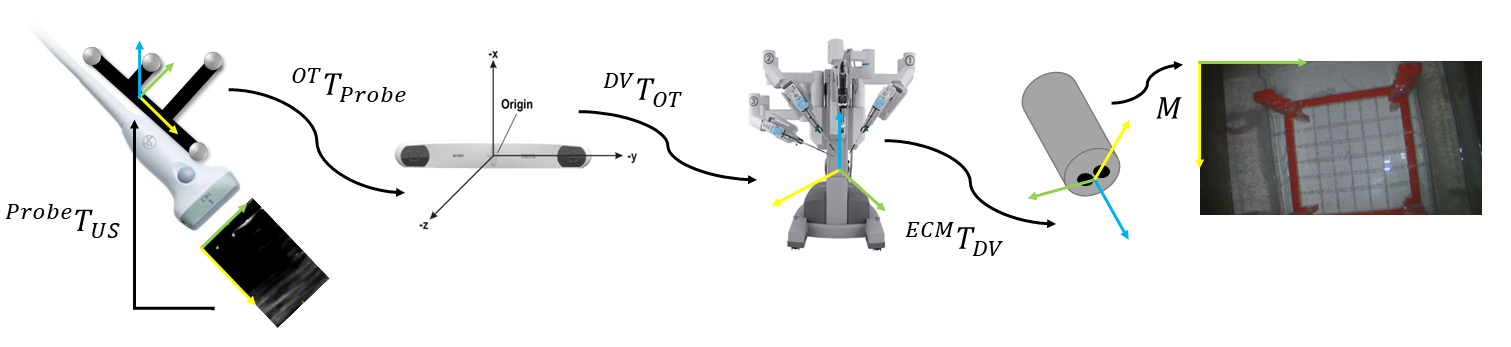}
            \caption{The hardware systems and the transformations in the system calibration to transform from the US image frame to the camera image frame.}
            \label{fig:system}
        \end{figure}

\begin{figure}
            \centering
            \includegraphics[width=0.8\textwidth]{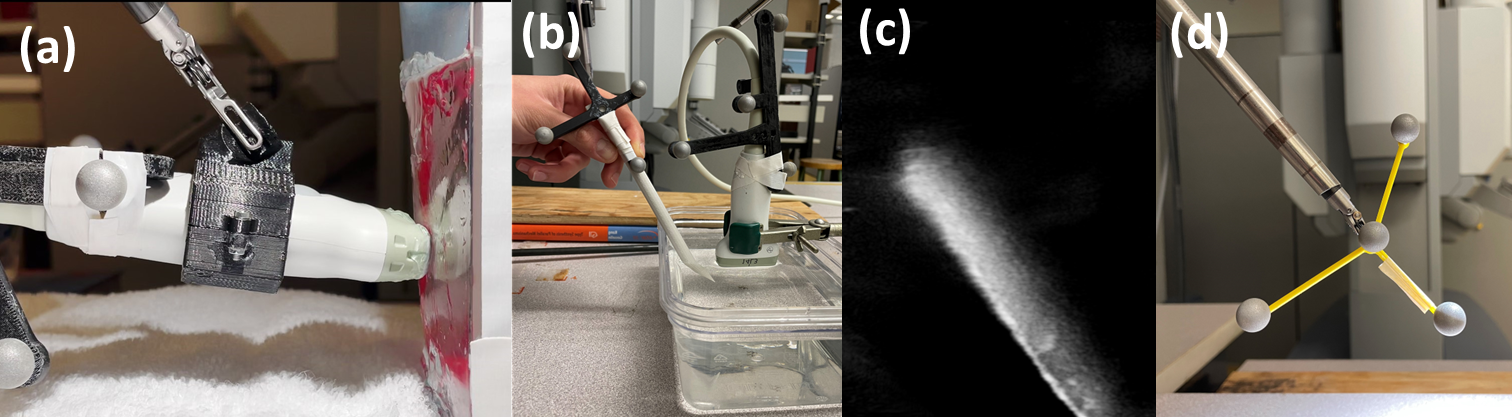}
            \caption{(a) 3D-printed holder grasping the US transducer; (b) tracked US transducer and stylus; (c) a US image of the stylus tip; (d) robot tool holding our designed marker for robot-tracker calibration. }
            \label{fig:us_calib}
        \end{figure}

        Fig.~\ref{fig:system} shows the components of our system. 
        Our surgical robot is the da Vinci~S surgical system (Intuitive Surgical, Sunnyvale, CA). 
        The US system is a BK3500 with a 14L3 linear 2D transducer (BK Medical, Burlington, MA). 
        A Polaris Spectra (Northern Digital, ON, Canada) tracks the US transducer.
        To enable US probe maneuvering by the surgeon using the da Vinci's third manipulating arm, we designed and 3D-printed a US transducer holder that can be grasped with the da Vinci ProGrasp in Fig.~\ref{fig:us_calib}.
        
    \subsection{System Calibration} \label{sec:system-calibration}
        Eq.~\ref{eq:transform} shows the transformations to project the US volume to the camera image ($^iT_{US}$). The endoscopic camera (ECM) location $^{DV}T_{ECM}$ can be read from the da Vinci API and the marker on the probe can be tracked by the tracker $^{OT}T_{Probe}$. Therefore, the system calibration needs to find the transformation from US image to the marker on the probe $^{Probe}T_{US}$, from optical tracker to robot $^{DV}T_{OT}$, and the camera projection matrix $M$. 
        \begin{equation}
            ^{i}T_{US} =M\ ^{ECM}T_{DV} \ ^{DV}T_{OT} \ ^{OT}T_{Probe}\ ^{Probe}T_{US} \label{eq:transform}
        \end{equation}
        \textbf{US to Optical Tracker Calibration:} 
            We used PLUS~\cite{Lasso2014a} for US-tracker calibration. A 3D-printed marker is attached to the US transducer to localize it in the tracker coordinate frame $OT$. 
            PLUS enables collecting time-synchronized transducer locations and US images. We first run the temporal calibration to correct the time offset, then we collect a tracked US video sequence-- imaging the tip of a tracked stylus pointer with the stylus tip and shaft visible in the US images. The transformation $^{Stylus}T_{tip}$ is known after a pivot calibration. We then use the fiducial registration in 3D Slicer to manually select the stylus tip location in the US images and to estimate a similarity transformation $^{Probe}T_{US}$ ($Probe$ is the coordinate frame of the US transducer marker).
            \[
               \ ^{Probe}T_{US} = {\arg\min}_T \|^{OT}T_{Stylus}\ ^{Stylus}T_{tip} - \ ^{OT}T_{Probe}T\ ^{US}T_{tip}\|_2
            \]
        \textbf{Robot to Optical Tracker Calibration:}
            We attach a customized marker with a known marker coordinate origin to the robot tooltip and collect the corresponding points in the tracker and robot coordinate frames. We estimate a rigid transformation between the corresponding points for $ ^{OT}T_{DV}$. 
            \[               
             ^{OT}T_{DV} = {\arg\min}_T \|^{OT}T_{marker} - T\ ^{DV}T_{Tip}\|_2
            \]
        \textbf{Camera Projection Matrix Calibration:}
            We use the camera calibration algorithm in~\cite{kalia2021preclinical} to estimate the camera projection matrix $M$, which includes the hand-eye calibration matrix (from ECM to the camera world coordinate system) and the camera intrinsic matrix. The algorithm detects key points on the robot tool in the robot and image coordinates to estimate $M$. 
            
    \subsection{MRI-3D US Registration}\label{sec:mri-us-registration}
        MRI-US registration is to find a spatial transformation  $^{MRI}T_{US}$. After the registration, we can map the MRI volume or the anatomy model generated from the MRI to the camera image by $^{i}T_{MRI} = \ ^{i}T_{US}\ ^{US}T_{MRI}$.
        To simplify the transformation in the AR system, we assume $^{MRI}T_{US}$ is an affine transformation. We first manually rigidly pre-register the 3D US volume and axial T1 based on the US transducer position and anatomy structures, and the Linear Correlation of Linear Combination (LC2) optimization~\cite{wein2013global} (ImFusion GmbH, Munich, Germany) is used to refine $^{US}T_{MRI}$. We evaluate the MRI-US registration for two different types of 3D US. We used a 3D US transducer (xMATRIX X6-1, Philips Healthcare, Bothell, WA) to acquire 3D US of the oropharynx in five healthy volunteers and four patients with oropharyngeal cancer.
        We then used the BK3500 14L3 linear transducer and the optical tracker in Section~\ref{sec:system-hardware} and PLUS to record the tracked US image sequences and reconstruct freehand 3D US for three healthy volunteers. We compound the 2D US images into a 3D volume using the US probe location provided by the optical tracker with the transformation  $^{Probe}T_{US}$  from Section \ref{sec:system-calibration} using PLUS, with linear interpolation and maximum compounding mode. We use the stick hole-filling with a length of 9 voxels.

    \subsection{Preoperative US-Intraoperative US  Registration}
        To evaluate the feasibility of preoperative to intraoperative 3D US deformable registration, we acquire freehand 3D US of two TORS patients, with the system and method of Section~\ref{sec:mri-us-registration}, before and after tongue suspension (retraction). 
        We minimize the mean square error between two volumes using QuickSyN ANTs~\cite{avants2009advanced}. We initialized the registration with a simple 3D translation to bring the US volumes roughly in the same space.

\section{Experiments and Results}\label{sec:experiments}

    \subsection{System Evaluation}
    
        To evaluate the accuracy of the robot-tracker calibration, we collected 100 pairs of points and used 50\% to estimate $^{DV}T_{OT}$ and 50\% for testing. The mean projection error is $1.73\pm0.86$ mm and $1.84\pm0.57$ mm for training and testing respectively.
        We measure the stylus-based calibration error by using 25 points to estimate $^{probe}T_{US}$ and 17 for testing, and the mean projection error is $0.59\pm 0.32$ mm and $0.78\pm0.41$ mm for training and testing respectively. 
        
\begin{figure}
            \centering
            \includegraphics[width=0.9\textwidth]{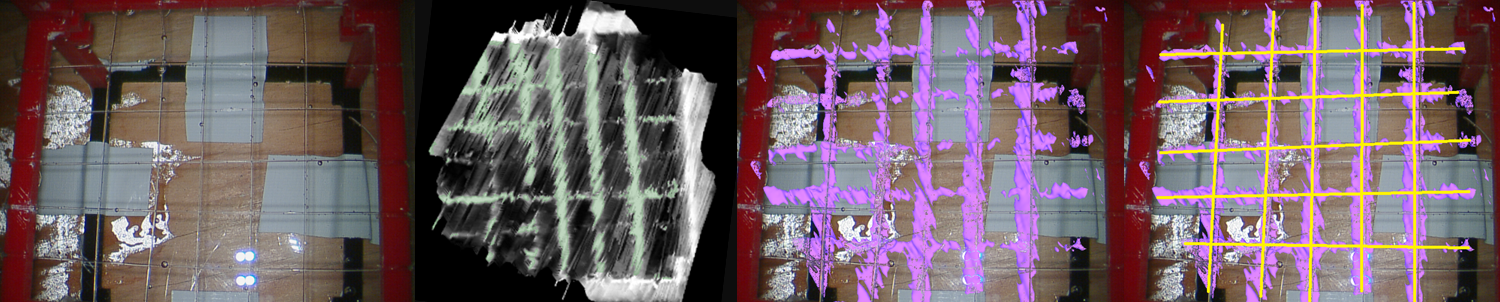}
            \caption{The water-bath experiment. From left to right: (1) original image; (2) 3D freehand US volume and detected wires; (3) projected wires in camera; (4) labeled wires and grids in the camera image.
            }
            \label{fig:grid_projection_exp}
        \end{figure}

\begin{figure}
        \centering
        \includegraphics[width=0.35\textwidth]{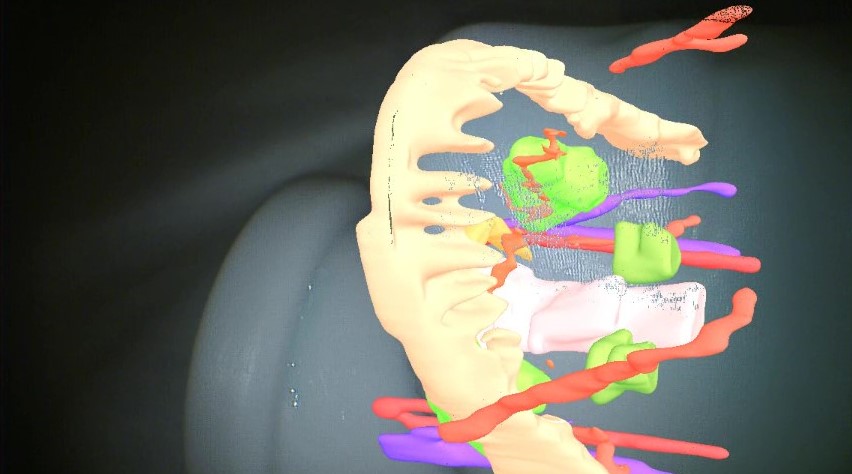}
        \includegraphics[width=0.35\textwidth]{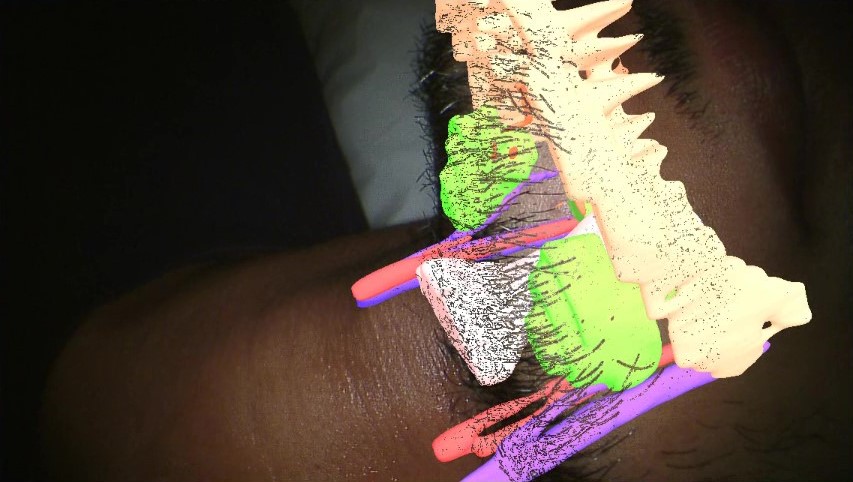}
        \caption{Images of the AR guidance system viewed in the camera frame.}
        \label{fig:AR_render}
    \end{figure}

        We  evaluated the system accuracy $^{i}T_{US}$ in a water-bath experiment in Fig.~\ref{fig:grid_projection_exp}. We used a 3D-printed structure and nylon wires to build 25 grid points, with nominal grid size of 10~mm, within the accuracy of the 3D printer. 
        We used freehand 3D US to image the grids and thresholded by intensity to segment the wires. We then projected the wires from the US to the camera and labeled the projected grid and the true grid locations on the camera image directly. We put the structure in three different locations and collected 75 data points. We also use triangulation~\cite{kalia2021preclinical} to estimate the location of 86 grid points projected from US and the ground truth in the camera world coordinate frame, and to compare their Euclidean distance as the projection error. The image size is $540\times960$, and the mean projection errors are in Table~\ref{tab:grid_projection_mse}. 
        The final system demonstration is in Fig.~\ref{fig:AR_render}, projecting the anatomy models segmented from the MRI to the camera with $^iT_{MRI}$ on a human subject. 
        
        \begin{table}[]
            \centering
            \caption{Mean projection error of the points projected from US to the camera.}
            \label{tab:grid_projection_mse}
            \begin{tabular}{|c|c|c|c|}
                \hline
                 & Left Camera & Right Camera & Camera World Frame\\
                 \hline
                 Projection Error & $27.14\pm9.78$  (pixel) & $26.03\pm7.95$  (pixel) & $9.93\pm6.65$ (mm)\\
                \hline
            \end{tabular}
        \end{table}
    
    \subsection{Feasibility Evaluation of MRI-US Registration}\label{sec:MRI-US-registration-results}
        
\begin{figure}
        \centering
        \includegraphics[width=0.47\textwidth]{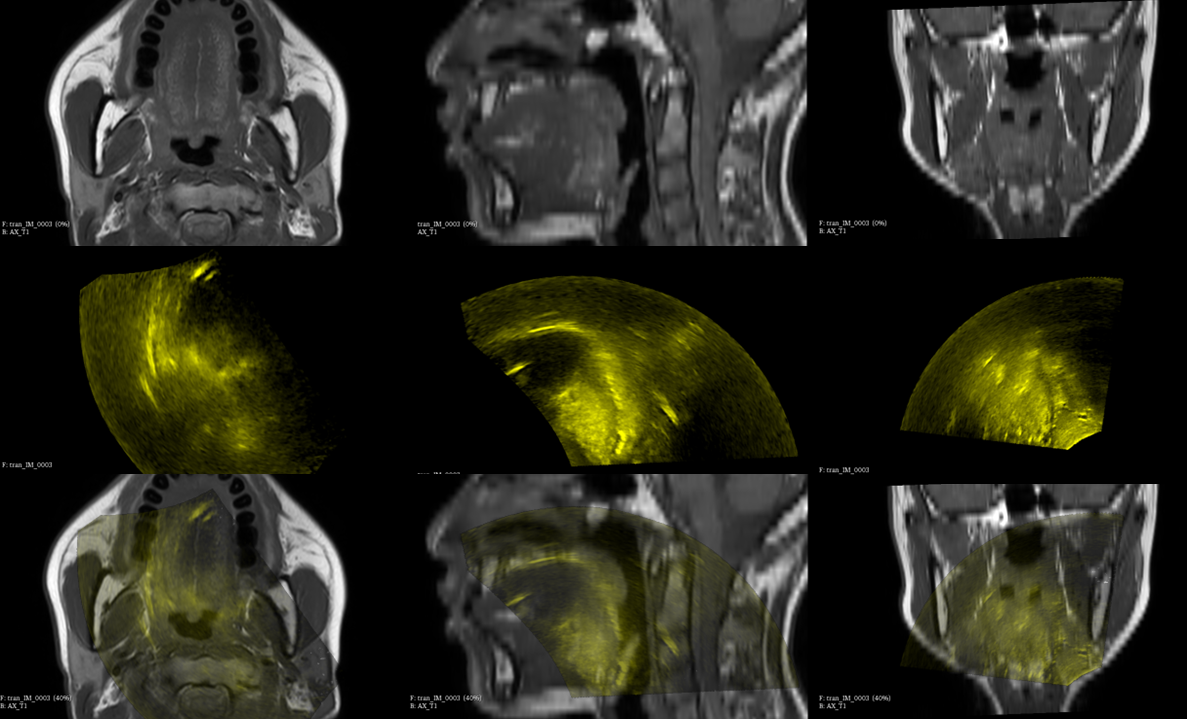}
        \includegraphics[width=0.46\textwidth]{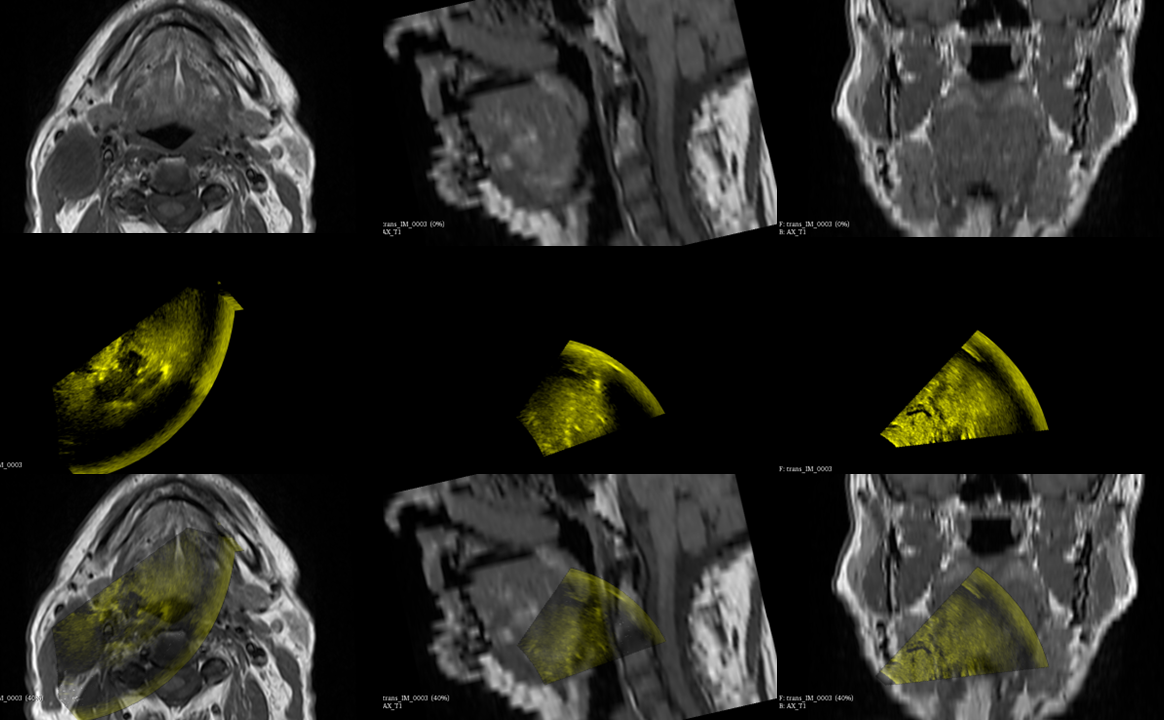}
        \caption{MRI-US registration using a 3D US probe \cite{chen2022Feasibility}.}
        \label{fig:philips_registration}
    \end{figure}

\begin{figure}
        \centering
        \includegraphics[width=0.8\textwidth]{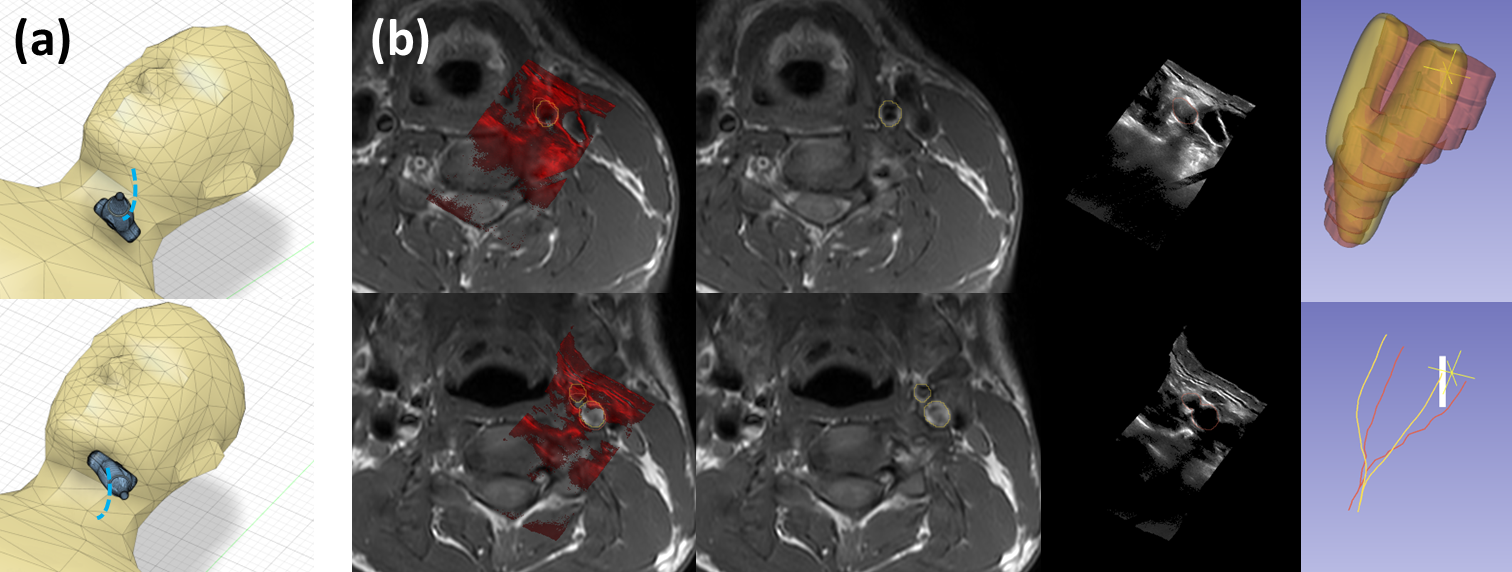}
        \caption{MRI-freehand 3D US registration. (a) scanning trajectory (models from \cite{Lasso2014a});
        (b) MRI-US overlay of the reconstructed 3D US, MRI, US, vessel segmentation and centerlines (yellow: MRI, red: US).}
        \label{fig:freehand_registration}
    \end{figure}

        MRI was acquired using a 3T Philips Elition MRI (Philips Healthcare, Best, Netherlands) for healthy volunteers, with axial resolution of 0.35 to 0.45~mm in the transverse plane and 3.6 to 4.5~mm in the axial direction. For patients, a Siemens Magnetom Espree 1.5T MRI (Siemens Healthcare, Erlangen, Germany) was used with a resolution of 0.375~mm in the transverse plane and 3.6~mm in the axial direction. The 3D probe-collected US details are in~\cite{chen2022Feasibility}.
        A radiologist (EP) manually selected anatomical landmarks to evaluate the target registration error (TRE). The TRE for MRI-3D probe-collected US is in Table~\ref{tab:TRE} and Fig.~\ref{fig:philips_registration} shows two examples. For freehand 3D US, the image depth is 6~cm at 9~MHz and the US volume spacing is 0.1~mm$^3$. We scanned the neck from the common carotid to the submandibular gland (Fig.~\ref{fig:freehand_registration}), segmented the carotids that appeared in both MRI and US and used Vascular Modeling Toolkit \cite{hahn1909integration} to extract centerlines. The TRE and the centerline distance\footnote{We interpolate each centerline into 1000 points and report the average distance between the corresponding points.} are in Table~\ref{tab:freehand_evaluation}, and Fig.~\ref{fig:freehand_registration} shows the example results. 
        
        \begin{table}[]
            \centering
            \caption{The registration error (TRE) for MRI-3D probe-collected 3D US \cite{chen2022Feasibility}.}
            \begin{tabular}{|c|c|c|c|c|c|c|}
            \hline
             Plane & \multicolumn{2}{c|}{Transverse} & \multicolumn{2}{c|}{Axial}      & \multicolumn{2}{c|}{Total}        \\
             \hline
            TRE  (mm)            & HVs & Patients  & HVs  & Patients  & HVs  & Patients  \\
            \hline
            By Group & 7.29±6.50          & 6.45±3.47 & 3.04±4.33          & 5.94±6.23 & 8.26±7.41          & 9.63±5.91 \\
            \hline
            All          & \multicolumn{2}{c|}{6.89±5.32}  & \multicolumn{2}{c|}{4.39±5.49}  & \multicolumn{2}{c|}{8.90±6.79} \\
            \hline
            \end{tabular}
            \label{tab:TRE}
        \end{table}

    \begin{table}[]
        \centering
        \caption{The target registration error (TRE) and the average distance of the vessel centerlines for MRI-freehand 3D US registration.}\label{tab:freehand_evaluation} 
\begin{tabular}{|c|cccc|c|}
\hline
                   & \multicolumn{4}{c|}{TRE (mm)}                                                                                                        &                                         \\
\multirow{-2}{*}{} & \multicolumn{1}{c|}{Transverse} & \multicolumn{1}{c|}{Axial}     & \multicolumn{1}{c|}{Total}     & Total (manual)               & \multirow{-2}{*}{Vessel Distance (mm)} \\ \hline
HV1                & \multicolumn{1}{c|}{3.78±0.47}  & \multicolumn{1}{c|}{5.40±1.80} & \multicolumn{1}{c|}{6.63±1.73} & 6.88±1.69                        & 2.15                                    \\ \hline
HV2                & \multicolumn{1}{c|}{3.18±0.84}  & \multicolumn{1}{c|}{0.89±1.56} & \multicolumn{1}{c|}{3.61±1.01} & 3.61±1.00 & 2.13                                    \\ \hline
HV3                & \multicolumn{1}{c|}{5.48±2.15}  & \multicolumn{1}{c|}{3.45±2.81} & \multicolumn{1}{c|}{7.33±0.86} & 7.56±0.87 & 2.67                                    \\ \hline
Average            & \multicolumn{1}{c|}{4.34±1.88}  & \multicolumn{1}{c|}{2.88±2.79} & \multicolumn{1}{c|}{5.85±2.05} & 6.00±2.13 & 2.32±0.25                               \\ \hline
\end{tabular}
    \end{table}
    
    \subsection{Feasibility Evaluation of US-US Registration}
\begin{figure}
        \centering
            \includegraphics[width=0.8\textwidth]{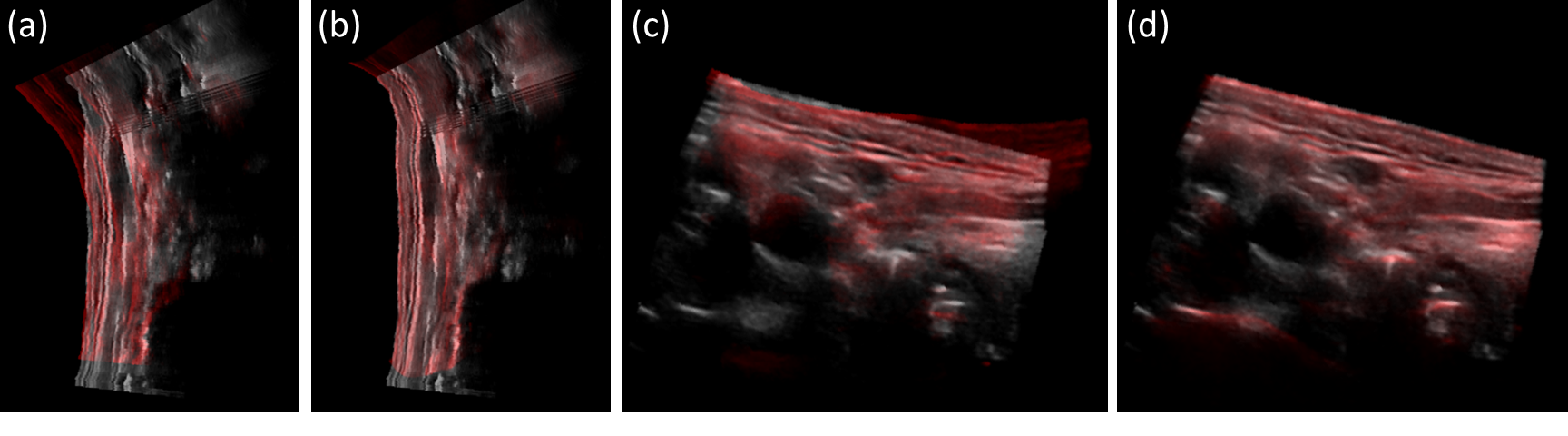}
        \caption{Pre-intra-operative deformable 3D US registration (grey: preoperative, red: intraoperative). (a) before registration (sagittal); (b) after registration (sagittal); (c) before registration (axial); (d) after registration (axial).
        }
        \label{fig:pre_intra_freehand_registration}
    \end{figure}

    The image depth is 4~cm at 9 MHz, and the reconstructed volume resolution is 0.2~mm$^3$. We used the same freehand scanning technique and evaluation in Section~\ref{sec:MRI-US-registration-results}. We also report the max deformation of the deformation field with registration evaluation in Table~\ref{tab:us-us-registration}, and Fig.~\ref{fig:pre_intra_freehand_registration} shows an example registration.

\begin{table}[]
    \centering
    \caption{The TRE, the average distance of the vessel centerlines, maximum deformation and run time for preoprative-intraoperative 3D US registration.}\label{tab:us-us-registration}
    \begin{tabular}{|c|c|c|c|c|}
    \hline
            & TRE  (mm)      & Vessel Distance (mm)       & Max Deformation (mm) & Time (s)\\ \hline
    P1      & 11.88±6.70 & 0.92      & 14.09      & 195.4
     \\ \hline
    P2      & 3.91±1.64  & 3.47      & 10.21     &  167.4
    \\ \hline
    Average & 7.90±6.30  & 2.19±1.27 & 12.15±1.94  & 181.4±14    \\ \hline
    \end{tabular}
\end{table}

\section{Discussion}

    In the system calibration evaluation,
    the experiment shows that the final system accuracy from the US to the camera is approximately 27 pixels (2.5\% of the image diagonal), similarly with~\cite{kalia2021preclinical} (26–30 pixels), even though we have an additional optical tracker to provide more degrees of freedom for the US transducer.
    The overall system error can be introduced by the calibration error, noise in the optical tracking, and grid labeling error. 
    US-tracker calibration can introduce a 0.8 mm error. Tracker-robot calibration is less than 2~mm, the camera calibration error is approximately 4.5~mm~\cite{kalia2021preclinical}, and our overall system error is 9.93~mm, which is comparable with the previously reported TORS guidance system error of 7.6~mm~\cite{ma2017intraoperative} for preoperative CT to tracker registration (only a qualitative evaluation is presented in~\cite{chan2020augmented}). A large part of the error is due to the calibration of the da Vinci S system camera. This will be reduced with the pre-calibrated camera of the da Vinci Xi system.

    We evaluate  MRI-US registration for two different 3D US systems. 
    Compared with~\cite{chen2022Feasibility},
     freehand 3D US with higher imaging frequency provides more detail in the near-surface and can scan a larger region. 
    The carotid and its branches are reliable landmarks for MRI-freehand 3D US registration. 
    The tissue deformation introduced by the forces applied to the participants during US scanning increases the registration difficulty, error and visual offset in the AR display. Fig.~\ref{fig:AR_render} shows that the anatomy displayed on the mannequin is more realistic than on the human because the mannequin has a non-deformable surface that is similar to MRI. Furthermore, the US-US registration shows that tongue suspension can cause large deformations (10-20~mm). The results demonstrate the importance of using real-time US to provide intraoperative information about deformation. Our study shows that intraoperative US-US registration can accommodate the deformation caused by tongue suspension. The maximum deformation is 10-15~mm, and the deformable registration error is 4-12 mm. The final system error will include both the system calibration error and the deformable registration error that needs future evaluation.

    A limitation in our MRI-US registration is that we need manual initialization, but similar AR systems for TORS also need surgeons to manually register preoperative models to the endoscope~\cite{chan2020augmented}, or manually select the registration fiducials~\cite{liu2015augmented}. 
    Our previous work has explored the visibility of landmarks~\cite{chen2022Feasibility} and future improvements will include detecting these landmarks automatically to pre-register the images~\cite{zeng2022learning}.
    Another limitation is that the marker must be visible to the optical tracker, and missing tracking data can worsen the quality of the compounded US volume. In the future, we will improve the spatial arrangement and marker design to improve tracking reliability. We will also evaluate other US calibration methods such as~\cite{chen2016guided} that do not require manual fiducial selection to improve the accuracy and system usability. 
    The limitations in US-US registration are that this preliminary study only has a small amount of data, and the deformable registration algorithm can not be run in near real-time. We will collect more data to investigate using modern algorithms to improve the speed and the quality of the registration.
    Another limitation is that surgeons do not have visibility of the US probe. However, the system is still usable if an assistant maneuvers the US probe. Our future work is to use an external camera view piped to the da Vinci TilePro in the console.

\section{Conclusion}
    We propose the first proof-of-concept for a transcervical US-guided AR system for TORS. We conduct a feasibility study of transcervical 3D US-MRI registration for two different types of 3D US and preoperative-intraoperative US registration, showing that 3D US has the potential to visualize the anatomy and to correct the distortion intraoperatively. We propose a new AR system for TORS and evaluate system accuracy. Our results show that 3D US transcervical imaging is a promising approach for image guidance in TORS. 

\bmhead{Acknowledgments}
    We thank the financial support from Natural Sciences and Engineering Research Council of Canada (NSERC) Discovery Grant and the Charles Laszlo Chair in Biomedical Engineering held by Dr. Salcudean. We thank David Black, Nicholas Rangga, and Angela Li for their help in CAD.

\section*{Declarations}
\textbf{Funding:} Natural Sciences and Engineering Research Council of Canada (NSERC) Discovery Grant and Charles Laszlo Chair in Biomedical Engineering held by Dr. Salcudean.
\textbf{Competing interests:} The authors have no conflicts of interest.
\textbf{Ethics:} Institutional ethics approval (\#H19-04025) was obtained for this study. Informed consent was obtained from all participants.

\bibliography{sn-bibliography}


\end{document}